\documentclass{WileyMSP-template}
\usepackage{parskip}
\usepackage{todonotes}
\usepackage{gensymb}
\usepackage{amssymb}
\usepackage{amsmath}
\usepackage{mathtools}
\usepackage{upgreek}
\usepackage{siunitx}
\usepackage{caption}
\usepackage{subcaption}
\newcommand{\etal}{\textit{et al. }}
\usepackage{color} % For highlighting

\begin{document}

% See the \addtolength command later in the file to balance the column lengths
% on the last page of the document

\pagestyle{fancy}
\rhead{\includegraphics[width=2.5cm]{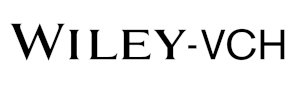}}

\title{Design of CLARI: A miniature modular origami passive shape-morphing robot}

\maketitle

% Author: Please give full first and last names for authors and include * after the name of all corresponding authors
\author{Heiko Kabutz,}
\author{Kaushik Jayaram $^{*}$}

% Dedication
% Optional dedication here. If no dedication is required, please leave blank
\dedication{}

% Affiliations: Please provide adacemic titles (Prof. or Dr.) for all authors where applicable, and include an institutional email address for all corresponding authors
\begin{affiliations}
Heiko Kabutz, Prof. Kaushik Jayaram\\
Paul M. Rady Department of Mechanical Engineering, University of Colorado Boulder, CO 80309.\\
Email Address: kaushik.jayaram@colorado.edu

\end{affiliations}

% Keywords: Please provide a minimum of three and a maximum of seven keywords, separated by commas

% Abstract should be written in the present tense and impersonal style (i.e., avoid we), and be at most 200 words long
\begin{abstract}
Miniature robots provide unprecedented access to confined environments and show promising potential for novel applications such as search-and-rescue and high-value asset inspection. The capability of body deformation further enhances the reachability of these small robots in complex cluttered terrains similar to those of insects and soft arthropods. Motivated by this concept, we present CLARI, an insect-scale 2.59g quadrupedal robot capable of body deformation with tethered electrical connections for power and control and manufactured using laminate fabrication and assembled using origami pop-up techniques. In order to enable locomotion in multiple shape configurations, we designed a novel body architecture comprising of modular, actuated leg mechanisms. Overall, CLARI has eight independently actuated degrees of freedom (two per modular leg unit) driven by custom piezoelectric actuators, making it mechanically dextrous. We characterize open-loop robot locomotion at multiple stride frequencies (\SI{1}{}-\SI{10}{\hertz}) using multiple gaits (trot, walk, etc.) in three different fixed body shapes (long, symmetric, wide) and illustrate the robot's capabilities. Finally, we demonstrate preliminary results of CLARI locomoting with a compliant body in open terrain and through a laterally constrained gap, a novel capability for legged robots. Our results represent the first step towards achieving effective cluttered terrain navigation with adaptable compliant robots in real-world environments.
\end{abstract}

\section{Motivation}
\label{sec:mot}

Today's robots are rapidly becoming highly capable due to tremendous recent innovations in design, fabrication, and control \cite{yang_grand_2018}. 
These mobile systems are increasingly successful when deployed in complex natural environments characterized by cluttered obstacles and confined spaces \cite{saranli_rhex_2001, spagna_distributed_2007, mongeau_rapid_2012, kalakrishnan_fast_2010, bjelonic_weaver_2018, faigl_adaptive_2019, bouman_autonomous_2020, lee_learning_2020}. 
In particular, miniature robots \cite{birkmeyer_dash_2009, jayaram_cockroaches_2016, haldane_robotic_2016, goldberg_gait_2017, jayaram_scaling_2020} provide unprecedented access to confined environments due to their small size and show promising potential for novel applications such as search-and-rescue \cite{murphy_disaster_2014} and high-value asset inspection \cite{de_rivaz_inverted_2018}. 

An analysis of the successful robots mentioned above highlights the importance of body geometry and mechanics as embodied intelligent solutions for achieving robust locomotion in complex natural environments. 
For example, robots that successfully navigate cluttered terrains without the need for sensing or active control leverage the tuned mechanics of their robust bodies to dissipate collisions with obstacles in their environment \cite{qian_dynamics_2015, jayaram_transition_2018, schiebel_mechanical_2019}.
These behaviors are often accompanied by changes to body orientation which are governed by the geometry \cite{li_terradynamically_2015, rieser_dynamics_2019, qian_obstacle_2020, han_shape-induced_2021} and mechanics \cite{jayaram_transition_2018, wang_directional_2020, astley_side-impact_2020, chang_anisotropic_2021} of the robot's body.
An alternate class of robots that successfully navigate confined terrains leverage the bioinspired strategy of exploiting the natural compliance of the appendages and bodies enabled by articulated geometries \cite{jayaram_cockroaches_2016, lathrop_towards_2021} or soft materials \cite{laschi_soft_2012, cianchetti_bioinspired_2015, tolley_michael_resilient_2014} to passively conform to environmental constraints. 
These studies have led to the resurgence of soft robotics as a leading paradigm for robot design in the last decade \cite{kim_soft_2013, laschi_soft_2016, rus_design_2015, rich_untethered_2018, hawkes_hard_2021}.

\begin{figure} [!htbp]
	%\RaggedLeft
    \centering
	\includegraphics[width=0.85\linewidth]{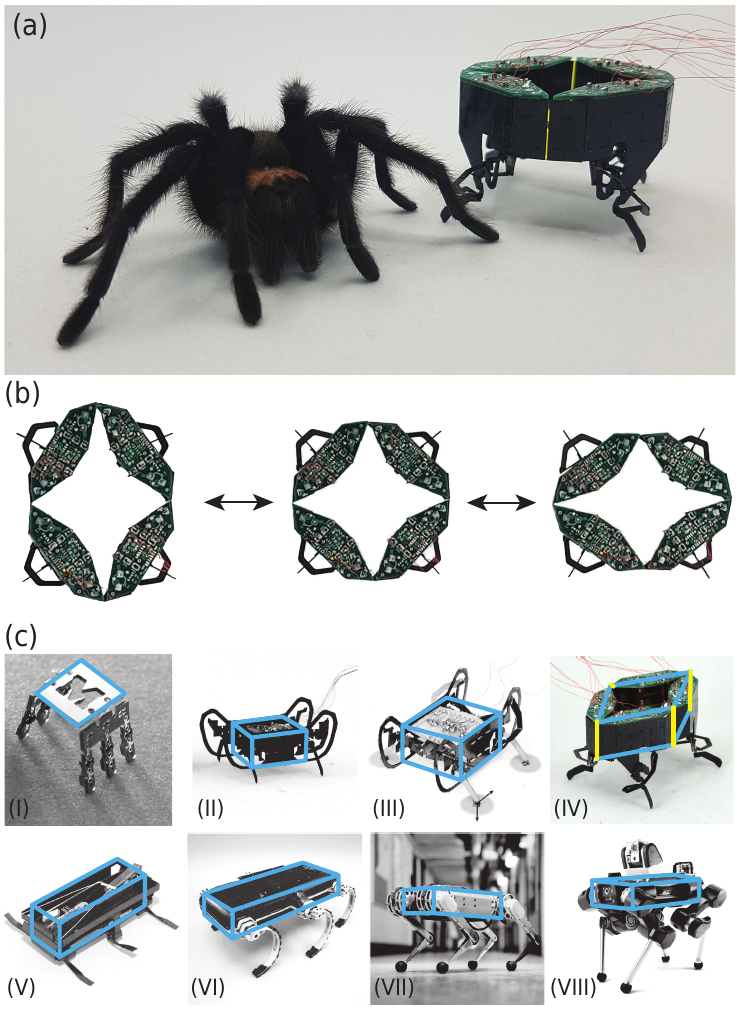}
	\caption{(a) CLARI --- Compliant Legged Articulated Robotic Insect, a miniature robot, featured next to an Oklohoma brown tarantula (\SI{\approx 30}{\milli\meter} body length) commonly found in Colorado. (b) CLARI's modular and compliant body allows it to vary body shapes and operate in multiple configurations. (c) Some of the most successful legged robots, ranging from \SI{}{\milli\meter} to \SI{}{\meter} sizes, all share a cuboidal body shape typically except for (IV) CLARI; (I)  Micro Robot \cite{vogtmann201725}, (II) HAMR-Jr \cite{jayaram_scaling_2020}, (III) HAMR-VI \cite{doshi_model_2015}, (IV) CLARI (this work), (V) DASH \cite{birkmeyer_dash_2009}, (VI) RHex \cite{saranli_rhex_2001}, (VII) MIT Cheetah \cite{seok_design_2015}, (VIII) ANYmal \cite{hutter_anymal_2017}.}
	\label{fig:Overall_spider}
\end{figure}

Despite this growing evidence, the majority of robots, across size scales, still maintain fixed body shapes (typically cuboidal, see Figure \ref{fig:Overall_spider}c, \cite{birkmeyer_dash_2009, goldberg_gait_2017, jayaram_scaling_2020, de_rivaz_inverted_2018, hoover_fast_2008, birkmeyer_clash_2011, haldane_animal-inspired_2013, goldberg_power_2018}) and are therefore unable to exploit the benefits of shape adaptation for complex terrain locomotion. 
%. 
One reason for this, especially on a smaller scale, is the increasing difficulty of design and fabrication associated with miniaturization \cite{st_pierre_toward_2019}. 
Another reason is the enhanced actuation and control efforts associated with a higher number of articulated degrees of freedom in the body \cite{de_croon_insect-inspired_2022}. 

As an initial step towards addressing these challenges, we present CLARI --- Compliant Legged Articulated Robotic Insect (Figure \ref{fig:Overall_spider}a) --- an insect scale quadrupedal soft tethered robot, the first in a series of articulated exoskeletal robots \cite{haldane_integrated_2015} potentially capable of body shape adaptation (Figure \ref{fig:Overall_spider}b). 
To motivate the design of CLARI, we specifically chose laterally confined environments such as gaps in between rocks, tunnels, or blades of stiff grass commonly found in natural terrains as our choice of complex terrain.
Unlike typical soft robots, which take advantage of material properties as the preferred solution to achieve body deformation, CLARI relies on an articulated morphology for lateral body compliance and shape change. 
This design of CLARI builds on our previous work with the robot CRAM (Compliant Robot with Articulated Mechanisms) which demonstrated cockroach-inspired dorsoventral body compliance during vertically confined space legged crawling \cite{jayaram_cockroaches_2016}. 
Our choice of articulated morphology for the robot body enables us to combine the rapid, autonomous, multi-gait locomotion capabilities of articulated laminate robots with the passive compliance-based embodied physical intelligence of soft material systems \cite{pfeifer_how_2006, miriyev_skills_2020, sitti_physical_2021, nguyen_adopting_2022}. 
Furthermore, we choose origami-based design \cite{sreetharan_monolithic_2012}, laminate fabrication \cite{wood_microrobot_2008} and the pop-up assembly process \cite{whitney_pop-up_2011} for CLARI as it offers an easy methodology of tuning geometry-based compliance (by varying flexure geometry) in addition to varying material properties (by changing individual layers) as needed within the laminate composite stacks of these robots at a variety of scales \cite{jayaram_scaling_2020}.

In this paper, we first describe the mechanical design principles of CLARI towards enabling cluttered terrain locomotion in Section \ref{sec:des_clari}. We follow this up by detailing the laminate fabrication technology used to manufacture the actuators, transmission linkages, and body of CLARI in Section \ref{sec:fab_exp}. We also describe the experimental procedures to characterize various robot subsystems and present our findings. Finally, we demonstrate unconfined robot locomotion in three fixed and one compliant body shape configurations across various frequencies and gaits and compare their performance in Section \ref{sec:rob}. Towards highlighting CLARI's potential as a soft robot capable of cluttered terrain navigation, we present initial evidence demonstrating it moving through a laterally varying gap utilizing body compliance. We conclude by summarizing our results and contrasting them with other state-of-the-art examples in the literature and outline the vision for future work with CLARI related to shape adaptation-enabled autonomous cluttered locomotion in Section \ref{sec:disc}. 

\section{Design of CLARI: Robot with Embodied Physical Intelligence}
\label{sec:des_clari}

\begin{figure} [!htbp]
	\centering
	\includegraphics[width=0.75\linewidth]{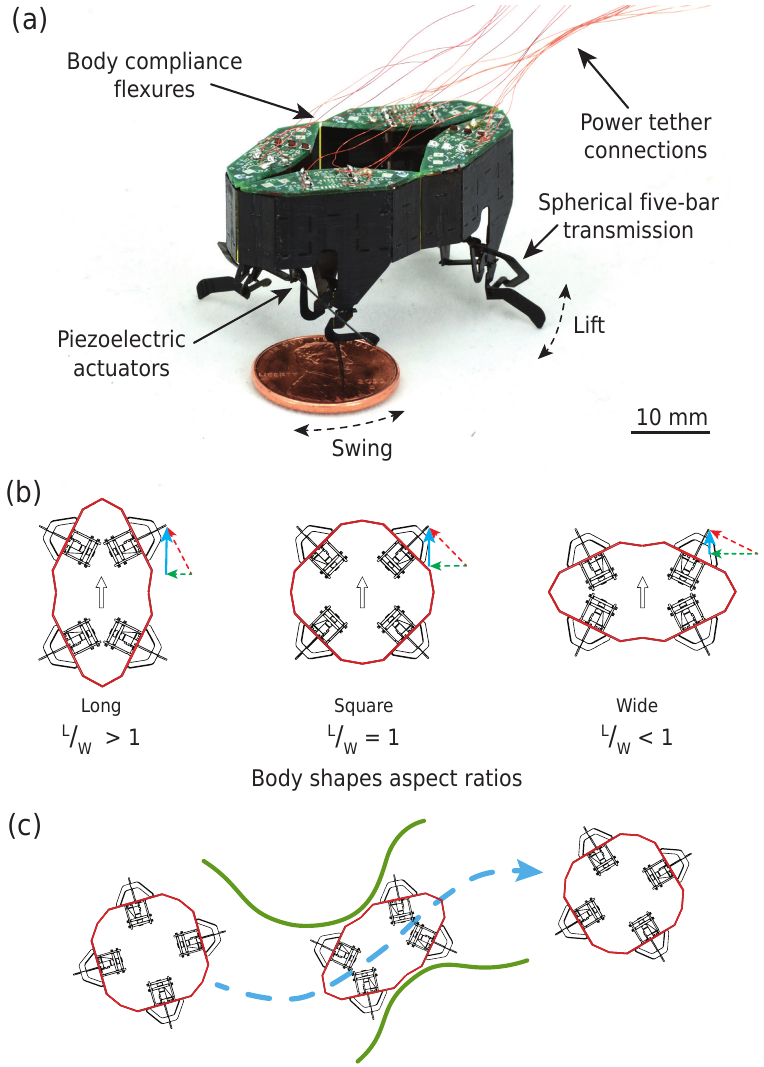}
	\caption{(a) A perspective view of CLARI highlighting the main features. (b) The three primary robot body shape configurations based on aspect ratios. The arrows show the ground reaction force vectors (red) with projections in the lateral (green) and propulsion (cyan) directions.(c) Artistic rendering of CLARI demonstrating laterally confined terrain navigation through a cluttered terrain leveraging the ability to adapt its body shape.}
	\label{fig:CompliantBodyShapes}
\end{figure}

In this paper, we introduce CLARI-1.0 (Figure \ref{fig:CompliantBodyShapes}), a 2.59g quadrupedal robot with eight independently actuated degrees of freedom (two per leg) driven by custom fabricated piezoelectric actuators and electrically tethered for power and control.
As a starting point for the design of CLARI, we leverage the Harvard Ambulatory MicroRobot (HAMR) series of insect-scale robots \cite{goldberg_gait_2017, jayaram_scaling_2020, doshi_model_2015, goldberg_high_2017}. 
To facilitate body adaptation, we introduce articulated body mechanisms in CLARI allowing the robot to be laterally compliant. 
While the actuators and transmission mechanisms in CLARI might seem similar to those in HAMR, we present a number of design and fabrication innovations in this work that are necessary to realize our robot. 
First, we introduce a modular single-leg design in CLARI to enable locomotion in multiple body shape configurations. 
This leg module integrates the necessary actuation, transmission, and power delivery mechanisms into a single unit, resulting in a decentralized (i.e., local) control architecture. 
Second, we align the actuators vertically within a CLARI leg module in order to achieve a compact leg design, unlike in HAMR, where they are horizontally oriented. 
This required a redesign of the spherical five-bar transmission mechanism, which further differentiates it from HAMR.
Finally, we implement the streamlined piezoelectric actuator fabrication process pioneered in Jafferis \etal \cite{jafferis_streamlined_2021}, a significantly improved and simplified workflow, which enables CLARI to be the first platform featuring such actuators. 
Each of these innovations is discussed in detail in the following sections.

\subsection{Compliant Body Design with Multiple Shape Configurations}
\label{sec:des_body}

Multiple body shape configurations of CLARI are potentially achievable due to body compliance enabled by the modular leg assembly design.
Each leg module is designed and fabricated as an individual unit and interconnected through single degree-of-freedom flexure joints to form a closed kinematic chain, i.e., a compliant body.
For the CLARI system, four individual leg modules are used, resulting in a rhomboid-shaped robot with variable aspect ratios (i.e., ratio of body length, \textit{L}, to body width, \textit{W}).  
Based on external constraints, the robot can passively deform and adapt to its environment (Figures \ref{fig:CompliantBodyShapes}b and \ref{fig:CompliantBodyShapes}c).
The stiffness between the individual leg modules can be tuned depending on the environment and ground surface roughness, independently or altogether.
In this paper, all intermodule joints are treated equivalently, resulting in a symmetric design for the first version of CLARI.

To characterize the effect of body shape on locomotion, we mechanically fixed the robot into three broad classes of body shape configurations --- \textit{long} (\textit{${\frac{L}{W}}>1$}), \textit{square} (\textit{${\frac{L}{W}}=1$}) and \textit{wide} (\textit{${\frac{L}{W}}<1$}) --- based on their aspect ratios (Figure \ref{fig:CompliantBodyShapes}b). 
In the \textit{long} configuration, the robot's legs are oriented favorably with respect to the body i.e., they swing primarily backward in order to propel the robot forward (see arrows in Figure \ref{fig:CompliantBodyShapes}b; the ground reaction force vector in red projects primarily onto the propulsion direction in blue and minimally onto the lateral direction in green). 
Therefore, we expect the highest locomotion performance (i.e., speed) in this configuration. 
The largest possible aspect ratio combination for the current version of CLARI is \textit{${\frac{L}{W}}=2.1$} with a \SI{44}{mm} body length and \SI{21}{mm} body width. 
In contrast, in the \textit{wide} configuration, CLARI's legs swing primarily laterally with respect to the body and orthogonal to the direction of forward locomotion (see the arrows in Figure \ref{fig:CompliantBodyShapes}b). 
As a result of this unfavorable leg positioning, we expect the robot to demonstrate low locomotion performance in this configuration. 
The smallest possible aspect ratio combination for the current version of CLARI is \textit{${\frac{L}{W}}=0.48$}.
Finally, in the \textit{square} configuration \textit{L=W=}$\SI{34}{mm}$ (\textit{${\frac{L}{W}}=1$}), CLARI is symmetric and we expect it to move with equal ease in both forward and lateral directions and thus potentially be capable of orthogonal locomotion.
However, since the legs are oriented at $\SI{45}{\degree}$ to either direction of locomotion, we expect forward performance to be inferior to the long configuration but superior to the wide configuration.

\begin{figure} [!htbp]
	\centering
	\includegraphics[width=\linewidth]{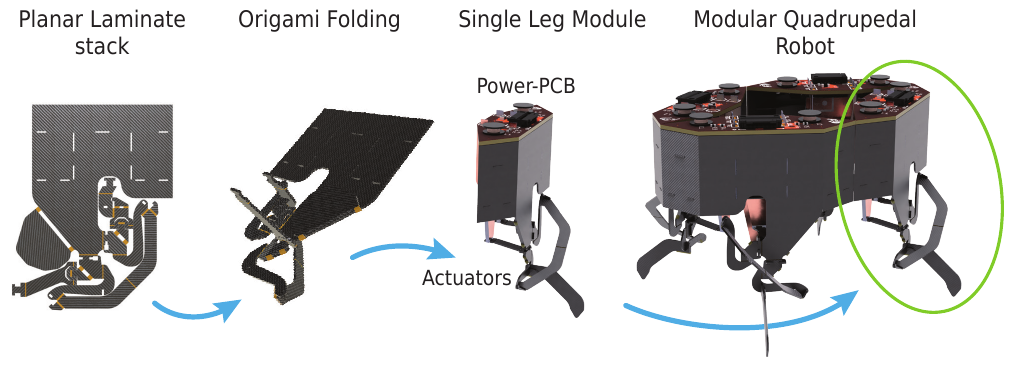}
	\caption{The different assembly steps involved in the CLARI fabrication. We start with planar laminate stacks of the SFB transmission which are folded into the correct orientation and integrated with actuators and the power-PCB to form a single-leg module. These individual leg modules are then assembled into the full robot.}
	\label{fig:ModularLegDesign}
\end{figure}

\subsection{Modular Leg Design}
\label{sec:des_modleg}

A single-leg module in CLARI includes two piezoelectric actuators, a spherical five-bar transmission mechanism to couple the degrees of freedom of elevation-depression, adduction-abduction, and protraction-retraction of the leg, and a printed circuit board (power-PCB) capable of housing power and sensing electronics (Figure \ref{fig:ModularLegDesign}). 
However, for all the experiments in this work, the power-PCB is used simply as an assembly component, which couples the actuators to the transmission on the robot body and relays power and control signals from an external computer via thin tether wires.
The power-pcb boards together add a payload of \SI{0.78}{g} to the robot (Table \ref{tab:MassDistribution}).

Modular leg design is key to achieving body compliance and shape adaptation in CLARI and thus separating it from other robots at this scale \cite{goldberg_gait_2017, jayaram_scaling_2020, doshi_model_2015, birkmeyer_dash_2009, goldberg_high_2017}. In addition to enabling a compliant body in CLARI, we note several further advantages of the modular leg design which are discussed in Section \ref{sec:disc}. 

\subsection{Transmission Design}
\label{sec:des_trans}

\begin{figure} [!htbp]
	\centering
	\includegraphics[width=0.55\linewidth]{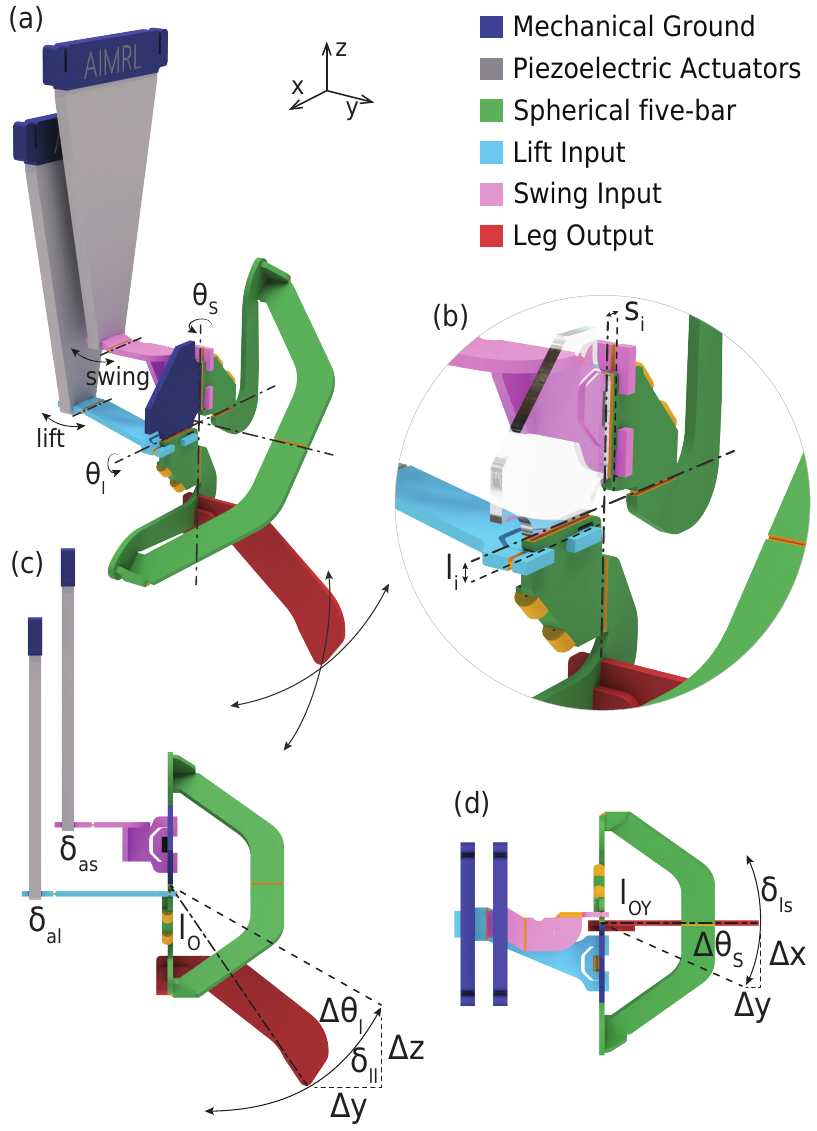}
	\caption{Overview of the CLARI leg mechanics featuring a spherical five bar (SFB) transmission. (a) A perspective view of the SFB with lift and swing actuators integrated. The relevant design parameters controlling the resulting leg trajectory are indicated and also summarized in Table \ref{tab:dim_clari}. (b) Close-up view of swing and lift linkages that amplify actuator motion. (c) Side projection view of leg module, indicating the change in y and z directions. (d) Top projection view of leg module, indicating the change in x and y direction.}
	\label{fig:Leg}
\end{figure}

Starting with the spherical 5-bar (SFB) linkage design inspired by the HAMR robots \cite{doshi_model_2015}, the leg linkage components in CLARI were redesigned to allow for vertical actuator placement.
This improved the overall compactness of the mechanism, facilitating assembly into an insect-scale robot while still permitting significant changes in aspect ratio for multiple body shape configurations.
To enable this modification, a number of small design changes were made to the SFB in CLARI relative to HAMR by adhering to fabrication and assembly considerations.
The most significant of these was the reduction in the total number of layers in the laminate from 11 to 5 by merging the transmission sublaminate and the chassis sublaminate into the same layers.
This modification used half the raw materials as before and reduced the linkage mass and the number of machining cycles.

The detailed design of our single-leg transmission is shown in Figure \ref{fig:Leg} with key sections color-coded by functionality. 
The mechanical ground (in blue) section of the transmission is part of the body frame and interlinks with the actuator mounting frame (i.e., power-PCB). 
Compared to earlier HAMR designs, these sections are doubly reinforced to minimize transmission losses.
The SFB origami structure (in green) when folded up cross-links the dynamics from the two actuators (lift, in cyan, and swing, in pink, respectively) through individual crank-slider mechanisms to the leg output (in red) glued to the lowest point of the SFB.
Both crank-slider mechanisms are actuated as close as possible (at a distance of \textit{$s_{i}$} and \textit{$l_{i}$} respectively for swing and lift, Figure \ref{fig:Leg}) to the center of rotation of the SFB to minimize off-axis bending, an issue with early generations of HAMR robots \cite{goldberg_gait_2017, doshi_phase_2017}.

The swing actuator primarily controls the protraction and retraction (hereafter referred to as swing motion) of the leg, which the lift actuator directly influences both elevation-depression (hereafter referred to as lift motion) and adduction-abduction (hereafter referred to as expansion motion) degrees of freedom (DoF) of the leg. 
In previous designs of HAMR and other small legged robots, the adduction-abduction motion has largely been ignored. 
During locomotion in open terrains, this DoF minimally influences overall locomotion performance, but could potentially become critical when navigating through laterally confined terrains. 
Although we acknowledge the adduction-abduction motion and characterize the same in Section \ref{sec:char_leg}, we did not specifically design for this DoF in the current version of CLARI and plan to address this in future iterations.
Thus, we adopt previously validated assumptions \cite{baisch_high_2014, doshi_model_2015} about SFB behavior in the quasi-static regime and model it as two separate single-input, single-output systems up until transmission resonant frequencies \cite{doshi_phase_2017}.
Using the procedure and convention in Doshi \etal \cite{doshi_model_2015}, we designed the leg transmission to amplify the displacement output of the lift and swing actuators (\textit{$\delta_{al}$} and \textit{$\delta_{as}$} respectively), while reducing the force output in each of the directions.
Our primary guiding principle was to have sufficient force output to support the robot weight on two legs in the vertical direction while still achieving stride lengths comparable to a similarly sized robot, HAMR6 \cite{goldberg_high_2017-1}.
To simplify our design, we chose the identical amplification joint distance \textit{$s_{i}$} and \textit{$l_{i}$} to result in the highest transmission ratios to meet the above criteria in swing ($T_s= l_{OY}/s_{i}$) and lift ($T_l= l_{OY}/l_{i}$).
The following table \ref{tab:dim_clari} summarizes the specifications of our chosen design parameters.

\begin{table}[!htbp]
    \centering
    \caption{Summary of CLARI design parameters}
    \begin{tabular}{c|c}
    \hline
        \textbf{Parameter} & \textbf{dimension}\\
        \hline
        \hline
        Actuator \\
        \hline
        Deflection amplitude, $\delta_{a} = \delta_{al} = \delta_{as}$ & $\pm$\SI{360}{\micro m} \\
        Blocked force, $F_{a} = F_{al} = F_{as}$ & $\pm$\SI{235}{mN} \\
        \hline
        \hline
        Transmission & \\
        \hline
        Lift input, $l_i$ & \SI{375}{\micro m}\\
        Swing input, $s_i$ & \SI{375}{\micro m}\\
        Lift transmission ratio, $T_l$ & $16$\\ %22.6
        Swing transmission ratio, $T_s$ & $16$\\
        \hline    
        \hline
        Leg & \\
        \hline
        Overall Length, $l_O$ & \SI{10.4}{\milli m}\\
        Lift output, $l_{OZ}$ & \SI{8.5}{\milli m}\\
        Swing output, $l_{OY}$ & \SI{6}{\milli m}\\
        Lift arc length, $\delta_{ll}$ & \SI{3.25}{\milli m} \\
        Swing arc length, $\delta_{ls}$ & \SI{2.95}{\milli m} \\
        Protraction - Retraction, $\Delta X$& \SI{2.85}{\milli m} \\
        Abduction - Adduction, $\Delta Y$& \SI{2.05}{\milli m} \\
        Elevation - Depression, $\Delta Z$& \SI{2.3}{\milli m} \\
        Vertical blocked force, $F_{leg}$ & $\pm$\SI{9.6}{\milli \newton} \\
        \hline
        \hline
        Body \\
        \hline
        Length (square)/Width (square) & \SI{34}{mm}\\
        Length (long)/ Width (wide) & \SI{44}{mm}\\
        Length (wide)/ Width (long) & \SI{21}{mm}\\
        Maximum Aspect Ratio (long) & 2.1\\
        Minimum Aspect Ratio (wide) & 0.48\\
        \hline
        \hline
    \end{tabular}
    \label{tab:dim_clari}
\end{table}

\section{Fabrication of CLARI and Characterization}
\label{sec:fab_exp}

In this section, we summarize the fabrication and characterization of the different components of CLARI. We start by presenting with a general overview of each of these steps and follow it up with component-specific details. 

\subsection{Fabrication Overview}
\label{sec:fab_sum}
To fabricate the components of CLARI including the chassis, transmission, actuators, flexure body links, etc., we adopt the planar laminate \cite{wood_microrobot_2008} kirigami manufacturing process (PC-MEMS,  \cite{sreetharan_monolithic_2012}) followed by pop-up folding assembly \cite{whitney_pop-up_2011}. 
In this paper, we used a custom-built (6D-Laser) femtosecond laser micromachine that allowed us to process a wider range of materials with higher fidelity compared to traditional nanosecond laser systems \cite{doshi_model_2015, jafferis_multilayer_2016}. 
The details of the individual fabrication steps for each of the CLARI components and any modifications from previous procedures are described in the Supplementary Information.

\subsection{Actuator Fabrication and Characterization}
\label{sec:char_act}

\begin{figure} [!htbp]
	\centering
	\includegraphics[width=0.45\linewidth]{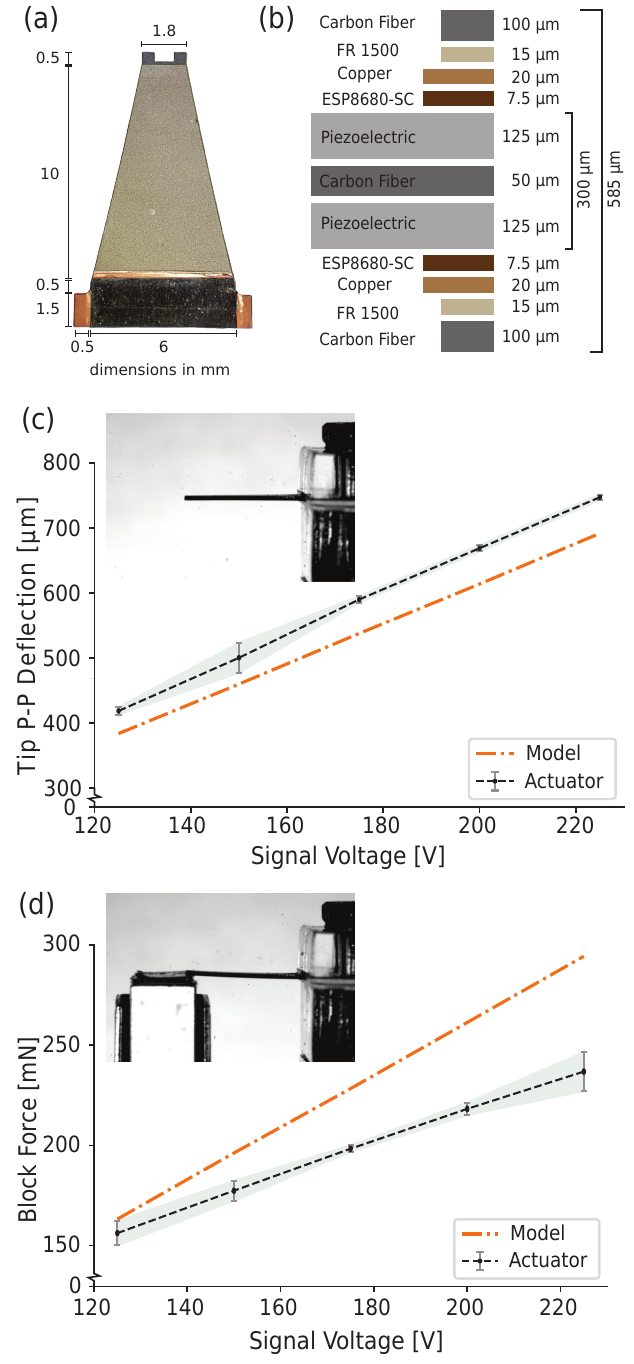}
	\caption{Design, fabrication, and performance of piezoelectric actuators. (a) Dimensions of the piezoelectric actuators in mm. (b) The stack of material layers used to create the bimorph actuator using the streamlined fabrication process \cite{jafferis_streamlined_2021}. (c) The plot of the peak-to-peak free tip deflection of the actuator with varying operating voltages. (d) The plot of the blocked force at the actuator tip with a fixed base at varying operating voltages.}
	\label{fig:Actuator}
\end{figure}

CLARI actuators were fabricated following the streamlined manufacturing process described by Jafferis \etal \cite{jafferis_streamlined_2021}, which overcomes the need for in-plane alignment of heterogeneous materials \cite{jafferis_multilayer_2016} and thus simplified fabrication and improved process yield and overall device performance.
Additional details are described in the Supplementary Information.
An example actuator fabricated using this process and its dimensions are shown in Figure \ref{fig:Actuator}a while the material stack-up is denoted in Figure \ref{fig:Actuator}b.
The operation of an example actuator is shown in the Supplementary Video.

The piezoelectric actuators were individually characterized for free tip displacement and tip block force performance at \SI{1}{\hertz} frequency and voltage steps of \SI{125}{\volt}, \SI{150}{\volt}, \SI{175}{\volt}, \SI{200}{\volt} and \SI{225}{\volt} using the setup (see the inset in Figure \ref{fig:Actuator}) described in the Supplementary information. 
At each bias voltage, five peak-to-peak sinusoidal cycles were powered, allowing for statistical analysis within specific cycles as well as between actuators at different voltages.
To predict the expected actuator performance, we used the actuator model from previous studies \cite{jafferis_design_2015, jayaram_concomitant_2018} and adapted it to our custom dimensions.

The free deflection measurements of the actuators indicate consistent performance (Figure \ref{fig:Actuator}c).
The measured free tip deflection of actuators is more than the model predictions and is consistent with the results published \cite{jafferis_design_2015, jafferis_streamlined_2021}.
Furthermore, at \SI{225}{\volt} the free tip peak-to-peak deflection is \SI{720}{\micro m} which is higher than previous measurements \cite{doshi_model_2015} in similarly sized actuators.
We suspect this is due to the minimal change in piezoelectric coefficients \cite{jafferis_design_2015} post-laser processing, an advantage of the cold ablation feature of femtosecond lasers \cite{pfeifenberger_use_2017} over nanosecond lasers (which were used for fabrication previously).  

Similarly, the blocked force measurements are consistent between the actuators (Figure \ref{fig:Actuator}d).
The observed results closely match the predictions of the model at \SI{125}{\volt}, and diverge to a performance drop of $17\%$ with respect to the model at \SI{225}{\volt}.
This is significantly closer to the model estimates than previous measurements, which indicated a \SI{32}{\%} difference at peak voltages \cite{jafferis_design_2015}.
However, the peak blocked force of the actuator at \SI{225}{\volt} is \SI{235}{\milli \newton}, which is lower than that measured previously for a similar sized actuator \cite{doshi_model_2015}.
We suspect this is possibly due to a slight movement of the actuators within the test jig providing a nonideal mechanical ground that will be improved in the future. 
We note that the addition of the carbon fiber reinforcement outer layers to the actuator was critical to maintaining rigidity at the base and improving force transmission. 
Without this structural addition, earlier versions of our actuators achieved only \SI{65}{\%} the best-recorded performance.

\subsection{Single Leg Module Fabrication and Characterization}
\label{sec:char_leg}
The leg modules are composed of the frame, transmission, actuator side walls, power-PCB, and leg in addition to the actuators described above and fabricated following the general procedures described for HAMR robots \cite{doshi_model_2015}. Additional fabrication details are described in the Supplementary Information.

The single leg modules were individually characterized for the displacement and the blocked force of the free leg tip at \SI{1}{\hertz} frequency and voltage steps of \SI{125}{\volt}, \SI{150}{\volt}, \SI{175}{\volt}, \SI{200}{\volt} and \SI{225}{\volt} in the lift and swing direction independently. 
At each bias voltage, five peak-to-peak sinusoidal cycles were powered, allowing for statistical analysis within specific cycles as well as between across leg modules at different voltages. 
The overall test setup (Figure \ref{fig:Leg_Experimental_Setup}) and the procedure used for these experiments is similar to the one described for actuator characterization in Section \ref{sec:char_act}.  
The operation of an example transmission is shown in the Supplementary Video.

\begin{figure*} [!htbp]
	\centering
	\includegraphics[width=0.95\linewidth]{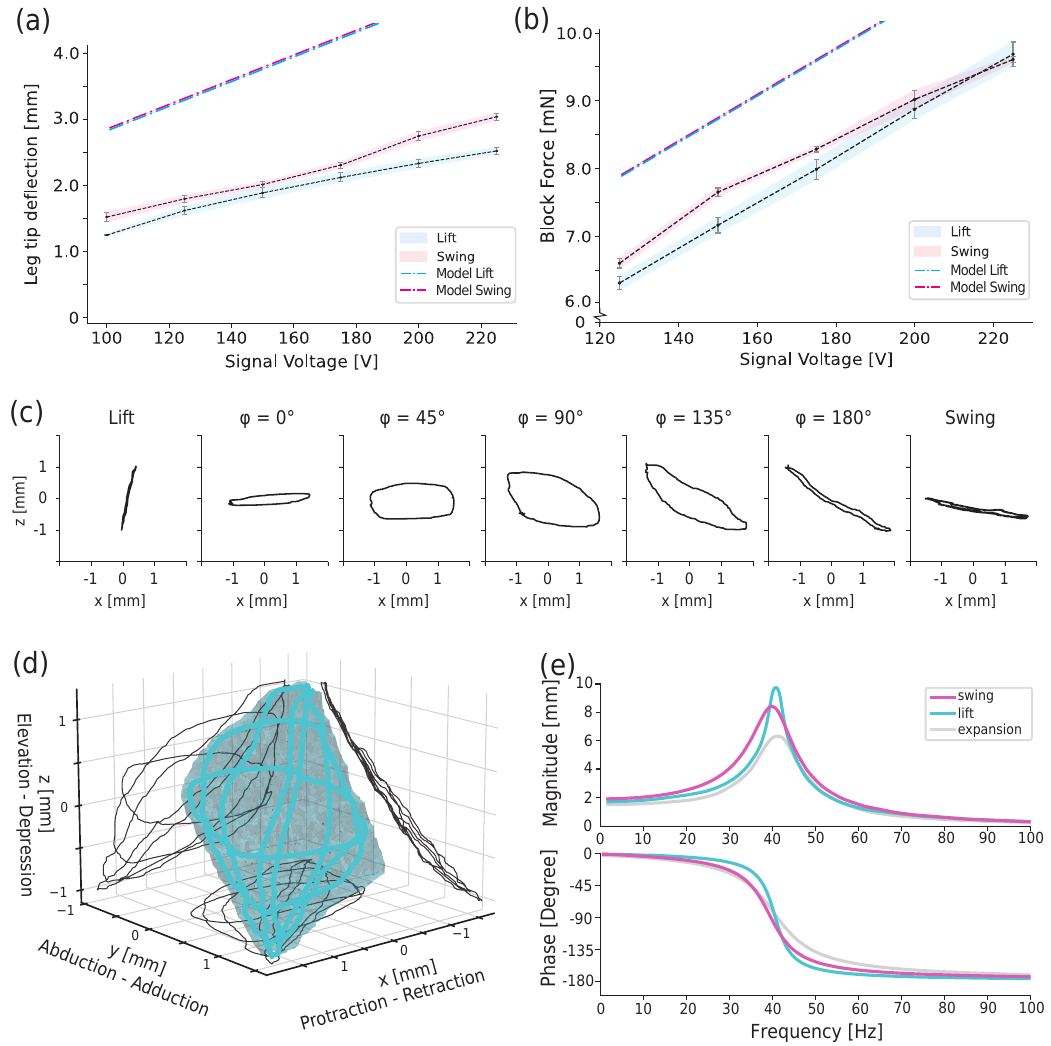}
	\caption{Single leg module characterization. (a) Peak-to-peak deflection of the CLARI leg tip at \SI{1}{\hertz} in the lift (blue) and swing (pink) directions. (b) CLARI leg tip blocked force in the lift (blue) and swing (pink) directions. Model predictions are included as dashed-dotted lines. (c) Projections of the leg tip onto the sagittal plane (xz) with varying lift and swing phase offsets reveal the diversity of foot trajectories. (d) The three-dimensional trajectory of the leg tip at varying intra-leg phase trajectories shows significant expansion motion, i.e. abduction-adduction, which is often ignored and might be potentially important, especially for laterally confined terrain locomotion. (e) The frequency response (bode plot) of leg motion to lift and swing signals reveal behavior analogous to second-order linear systems in respective output directions.}
	\label{fig:Leg_char}
\end{figure*}

Under our test conditions, we observed that the leg modules (averaged data across modules) showed near linear relationships with voltage for both leg tip deflection shown in Figure \ref{fig:Leg_char}a and blocked force Figure \ref{fig:Leg_char}b in both the lift (cyan) and swing (pink) directions.
The maximum measured blocked force in both swing and lift was about $9.6mN$ at \SI{225}{\volt}, resulting in an effective force transmission ratio of $T_F = F_{actuator}/F_{leg} \approx 235/9.6 = 24.5$ (which is less efficient than the modeled transmission ratio, $T_l=16$) indicating transmission efficiency of \SI{65}{\%}. 
With a length of $l_O=10.4mm$, the leg generated maximum displacements of about \SI{2.85}{mm} in swing and \SI{2.3}{mm} in vertical lift directions at 225V. 
These results are lower than those reported in similarly sized HAMR \cite{doshi_model_2015, doshi_effective_2019} and can be attributed to the higher transmission ratio used in CLARI.
However, our experimental results are well below the prediction of the model (Figures \ref{fig:Leg_char} a and b), indicating the need for significant improvements in fabrication and assembly to increase transmission efficiencies in future CLARI iterations. 
Possible sources of transmission losses include (but are not limited to) inelastic deformation of the flexures in addition to deformation in the linkages, imperfect mechanical ground for the actuator's connections, and degradation due to fatigue \cite{doshi_model_2015, doshi_effective_2019}.

Next, we demonstrate the ability to generate a variety of leg trajectory shapes in the quasistatic regime by varying the relative phase between the lift and swing actuation.
We depict their projections onto the robot's sagittal plane in Figure \ref{fig:Leg_char}c.
The experimentally measured trajectories were shaped as distorted ellipsoids varying from flattened near horizontal to near vertical and everything in between.
We suspect that these distortions were a result of non-ideal transmission assembly and are one of the potential improvements for future generations of CLARI.
For all the locomotion studies that follow in the remainder of the manuscript, we used a leg trajectory where the phase offset between lift and swing was \SI{90}{\degree} as the nominal trajectory shape.
We plotted the 3D leg reachability space in Figure \ref{fig:Leg_char}d and observed significant movement outside the sagittal plane in the abduction and adduction directions that could potentially enhance omnidirectional locomotion capabilities.
This observation suggests that a majority of coupled transmissions like SFB result in complex motions that potentially need further attention for their influence on locomotion but are often ignored. For example, \cite{doshi_model_2015, de_rivaz_inverted_2018, chen_inverted_2020} consider only 2D projections of leg trajectory shapes.

Finally, we also characterized the transmission dynamics by performing frequency sweeps from \SI{1}{\hertz} to \SI{100}{\hertz} at \SI{150}{\volt} and determined their Bode plot in Figure \ref{fig:Leg_char}e. 
We observed that the transmission behaved like a linear second-order under-damped system in each of the three measured directions. 
We quantified the resonance frequency in the swing and lift (and expansion) directions as \SI{40}{\hertz} and \SI{41.5}{\hertz} respectively and therefore expect the range of ideal running frequencies for CLARI to be up to \SI{40}{\hertz}.

\subsection{Whole Robot Assembly}
With individual leg modules characterized, four equally performing ones were selected and then connected through the side walls interlocking to horizontal ribs within each module, resulting in the complete robot (Figure \ref{fig:CompliantBodyShapes}). 
Additional fabrication details are described in the Supplementary Information.
The overall mass distribution for CLARI is summarized in Table \ref{tab:MassDistribution}.

\section{Locomotion Performance of CLARI}
\label{sec:rob}

In this section, we determine the locomotion of CLARI to highlight the effect of body compliance and shape on performance. 
As noted previously (Section \ref{sec:des_modleg}), the current version of CLARI is electrically tethered for receiving power and control signals from an off-board computer.
We follow the same procedure to drive the robot as we did for the actuators (Section \ref{sec:char_act}) and the single-leg modules (Section \ref{sec:char_leg}).
To quantify performance, we ran the robot with multiple gaits at stride frequencies of \SI{1}{}, \SI{5}{}, and \SI{10}{\hertz} and measured forward speed as a function of body compliance or shape.
All tests were performed at a fixed operating voltage of \SI{200}{\volt} in an open terrain arena covered with cardstock as the running surface and filmed at high speed (\SI{240}{fps}, side and top view synchronized) for post-processing and kinematic analyses.
Our key findings are described in the following.

\textbf{Gait Flexibility.}
CLARI is mechanically highly dexterous due to the eight independently controlled DoFs (two per leg module) and therefore able to operate in a variety of biologically inspired gaits including walk, trot, pronk, bound, and pace (see \cite{goldberg_gait_2017} for definitions). 
To demonstrate the various gaits, we mounted CLARI on a custom stand (legs not in contact with the ground) and varied the relative phasing of the actuators between the leg modules following the procedure detailed in \cite{goldberg_gait_2017} (see Supplementary video).
Although multiple gaits are feasible, we chose to focus on trot and walk for further locomotion characterization as they are the most common ones used by robots on this scale \cite{goldberg_gait_2017, jayaram_scaling_2020, goldberg_high_2017-1, doshi_effective_2019}. 
The gait timing diagram for these two gaits is shown in Figure \ref{fig:Leg_trajectories}. 

\begin{figure*} [!htbp]
	\centering
	\includegraphics[width=\linewidth]{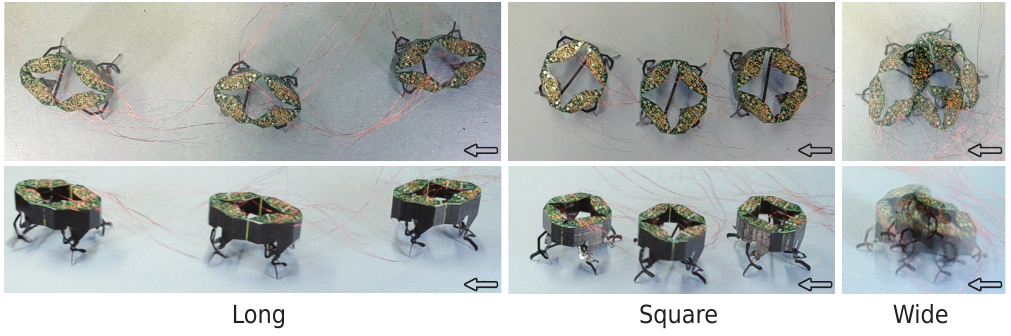}
	\caption{Robot locomotion (trot, \SI{10}{\hertz}) at different body shapes shown as time consistent snapshots.}
	\label{fig:CLARI_walking}
\end{figure*}

On the ground, with equal voltages applied to each leg actuator during a gait, we achieved near-straight line locomotion as shown in Figure \ref{fig:CLARI_walking}. 
However, by varying the applied voltages between actuators following the procedure in Goldberg \etal \cite{goldberg_high_2017}, we could realize turning with the same gaits although we did not extensively characterize it.
An alternate approach for turning in CLARI is by varying the relative phase between legs \cite{doshi_phase_2017}, but is not demonstrated here.
An example trial featuring a \SI{90}{\degree} turn is depicted in Figure \ref{fig:CLARI_turning} and the Supplementary Video. 

\textbf{Effect of Body Compliance.} Without specifically tuning the body joints between the leg modules, our design choices resulted in a highly compliant robot body. 
Once this robot was placed on the ground, we observed large body oscillations modifying the robot shape, which clearly highlighted CLARI's body compliance and the potential for laterally confined locomotion through shape-morphing. 
However, for any of the gaits, we did not observe any notable forward locomotion indicating that the body was too compliant and did not generate effective ground reaction forces for propulsion (see the Supplementary Video). 
By tuning the stiffness between adjacent leg modules \textit{ad hoc} using polyimide strips (\SI{12.5}{\micro \meter} thick, \SI{16}{\milli \meter} length and \SI{4}{\milli \meter} height; see yellow strips in Figure \ref{fig:CLARI_compress}), we were able to obtain significant forward motion (see Supplementary video) as quantified in Figure \ref{fig:CLARI_Speed_Gait}. 

\textbf{Effect of Body Shape.} 
To isolate the effect of body shape and quantify locomotion performance independent of compliance, we chose to fix the body (by gluing the opposite leg modules using thin carbon fiber rods and thus eliminating any body compliance) in three specific shapes (${\frac{L}{W}}$= 0.48, 1 or 2.1, i.e., the extremes in \textit{wide}, \textit{square} and \textit{long} classes described in Section \ref{sec:des_body}) and characterized locomotion performance in those configurations.
We measured the fastest and slowest forward running speeds with CLARI in the \textit{long} and \textit{wide} body configurations and intermediate speeds in the \textit{square} body shape (see Supplementary Video). 
Figure \ref{fig:CLARI_walking} visually illustrates these findings for the three different shape configurations tested and validates our hypotheses about the influence of body shape on locomotion due to the varying leg orientation for thrust production (Section \ref{sec:des_body}).
Furthermore, we deduce that in the \textit{square} configuration, CLARI moved with equal ease in the forward and lateral directions, indicating the potential for omnidirectional locomotion.

\textbf{Effect of Gait and Stride Frequency.} 
The running speed results from the different experiments are shown in Figure \ref{fig:CLARI_Speed_Gait}.

\begin{figure} [!htbp]
	\centering
	\includegraphics[width=0.75\linewidth]{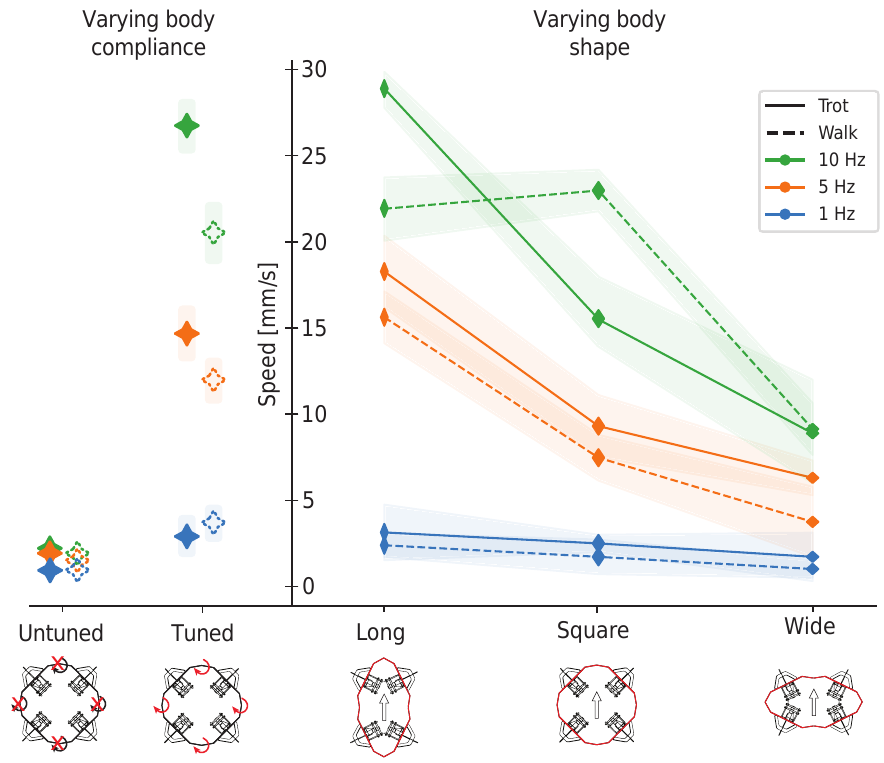}
	\caption{Robot locomotion speed performance as a function of body compliance and body shape at varying leg frequencies and varying gaits. Data is represented as mean $\pm$ 1 standard deviation. We find that the CLARI generally records the best performance in the long shape configuration matching our expectations.}
	\label{fig:CLARI_Speed_Gait}
\end{figure}

In general, we found that both gaits performed similarly, with the trot performing better than the walk at a given stride frequency and body shape except at \SI{5}{\hertz} in \textit{square} configuration. 
We posit that the timing of ground contacts during trotting enabled CLARI to take advantage of favorable body dynamics more than during walking, contributing to faster locomotion \cite{dickinson_how_2000}.
We measured the best locomotion performance with CLARI in \textit{long} body configuration with the highest speed of \SI{28}{\milli\meter\second^{-1}} at \SI{10}{\hertz}, which was comparable to that observed for HAMR \cite{doshi_effective_2019} for a similar stride frequency.
With a tuned compliant body, we measured CLARI's performance to vary between that in long and square body fixed shape configurations with trot gaits increasingly faster at higher frequencies.
Regardless of body shape, at stride frequencies higher than \SI{10}{\hertz}, we observed a strong detrimental influence of destabilizing dynamics that results in poor locomotion, which was analogous to those observed in other small robots in the body dynamics regime \cite{goldberg_gait_2017}.

\textbf{Towards Cluttered Terrain Locomotion. }
Having demonstrated the effect of body compliance and shape on locomotion performance in open terrain, we present initial evidence of the robot's ability for confined locomotion by demonstrating it moving through a laterally varying gap utilizing body compliance (see Supplementary Video). 
Details of the experimental setup are included in the supplementary text. 
Figure \ref{fig:CLARI_compress} shows that CLARI was able to passively deform by over \SI{30}{\%} laterally and vary its shape from an aspect ratio of 1.2 to \SI{1.9}{} and adapt to its environmental constraint to fit through the lateral gap of \SI{22}{\milli \meter}.
This demonstration is the first step towards CLARI being able to navigate effectively through various kinds of complex environments including cluttered and confined terrains. 

\begin{figure} [!htbp]
	\centering
	\includegraphics[width=0.5\linewidth]{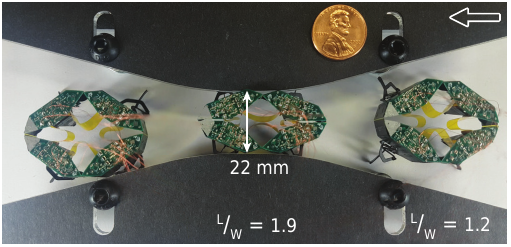}
	\caption{Robot compressing laterally while walking through a body constraining gap. The lateral walls interact only with the body and therefore do not limit leg movements.}
	\label{fig:CLARI_compress}
\end{figure}

\section{Discussion and Future work}
\label{sec:disc}

In conclusion, we successfully designed and fabricated CLARI, the first in a series of miniature legged robots with articulated modular bodies. 
Body compliance and shape change were achieved by interlinking four novel modular leg units in a closed kinematic chain. 
The electrically tethered CLARI was demonstrated to successfully locomote in multiple static body shape configurations lacking compliance at various gaits and stride frequencies. 
Similarly, the robot was also demonstrated to run effectively with an experimentally tuned compliant body at similar conditions. 
In addition to forward locomotion, CLARI was also able to execute turns on demand using the same gaits by varying the stride amplitude. 
In general, we found that the robot's running speed is comparable to other systems of this size at equivalent operating conditions \cite{goldberg_gait_2017, doshi_effective_2019}. 
More importantly, our experiments showed that body compliance and shape have significant effects on robot performance supporting our hypotheses about their contributions towards embodied physical intelligence and that they need to be tuned appropriately for effective locomotion in performance-specific applications.
Finally, we presented initial evidence of CLARI's ability for passive shape adaptation with a soft body of tuned compliance enabling laterally confined locomotion - a first for legged robots to the best of our knowledge (see Table \ref{tab:RobotComparison}). 
Thus, in this work, we have taken the first step towards autonomous cluttered terrain navigation by establishing the CLARI's ability to locomote in various shapes and with tuned compliance. 

\begin{table}[!htbp]
    \centering
    \caption{Comparing the shape morphing and complex terrain locomotion abilities of some of the successful legged platforms across sizes from a few millimeters to meters}
    \begin{tabular}{l|c|c|c|c}
    \hline
        \textbf{Robot} & \textbf{Length}  & \textbf{Aspect}  & \textbf{Shape} & \textbf{Complex} \\
         & \textbf{(cm)} & \textbf{Ratio} & \textbf{Change} & \textbf{Terrain} \\
        \hline
        Microrobot \cite{vogtmann201725} & 0.4 & 1 & No & Lab \\
        HAMR-Jr \cite{jayaram_scaling_2020} & 2 & 1.25 & No & Lab \\
        \textit{CLARI}  & \textit{3.4} & \textit{0.48-2.1}  & \textit{Yes} & \textit{Lateral}\\
        HAMR \cite{baisch_high_2014} & 4.4 & 1.25 & No & Lab \\
        DASH \cite{birkmeyer_dash_2009} & 13 & 2 & No &  Lab\\
        RHex \cite{saranli_rhex_2001} & 50 & 2.2 & No$^{*}$ &  Open$^{\dagger}$\\
        Cheetah \cite{seok_design_2015} & 60 & 1.7 & No$^{*}$ & Lab\\
        ANYmal \cite{hutter_anymal_2017} & 80 & 1.35 & No & Open$^{\dagger}$\\
        \hline
    \end{tabular}
    \label{tab:RobotComparison}
    \RaggedRight\\
    $^{*}$ Robot versions with spine/ backbone morphology have been explored \\
    $^{\dagger}$ Natural terrain with less than body height and unconstraining obstacles
\end{table}

A key innovation of this work is the modular leg assembly, atypical for robots at this scale. In addition to enabling a compliant body in CLARI, we noted several additional advantages of the modular leg design. 
This choice allowed us to build leg modules without worrying about their actual position on the robot which sped up the iteration cycles for optimizing fabrication by limiting our scope to a single subsystem. 
The modular approach also simplified the overall robot design process and potentially makes it easier to scale up to multi-legged systems with a high number of legs (e.g. centipede-like robots \cite{hoffman_turning_2012, ozkan_self_2021}). 
More importantly, self-contained modular legs allow for convenient repair and replacement of degrading individual appendages, and thus make the platform significantly easier to maintain relative to HAMR or other monolithic robots.
Furthermore, each individual leg module can be thoroughly characterized before assembly and matched with similarly performing sets and, therefore, minimizes the need for "trimming" \cite{dhingra_device_2020}, an essential process for effective open-loop locomotion such as running in, straight path. 
Overall, we believe that such modularity allows for easier and faster novel robot platform development with different leg arrangements, leg numbers, or body shapes, without having to redevelop and characterize entirely new leg designs.
However, we also acknowledge that numerous improvements are required before CLARI can fully achieve its goal of traversing complex terrains effectively, and we envision addressing them in future iterations of the robot. 
Some immediate next steps are discussed below.

CLARI is currently unable to move effectively without careful body-compliance tuning.
This is not surprising because the role of body mechanics during legged locomotion as a morphological computation principle \cite{pfeifer_how_2006, nguyen_adopting_2022} is an area of active interest to biologists \cite{alexander_principles_2003, wilson_locomotion_2013}, physicists \cite{chong_coordination_2021, chong_coordinating_2022}, and roboticists \cite{seok_design_2015, haynes_laboratory_2012, caporale_coronal_2020} alike and remains a challenging problem yet to be fully understood \cite{hawkes_hard_2021, sitti_physical_2021}. 
For example, the first couple of generations of the MIT Cheetah \cite{seok_design_2013} featured a compliant backbone which was later abandoned \cite{bledt_mit_2018} in part due to the challenges associated with tuning for effective locomotion.
Future generations of CLARI will explore a number of strategies including the systematic passive \cite{haynes_laboratory_2012} tuning of articulated flexures, the addition of actuated joints \cite{kellaris2021spider}, the integration of reconfigurable \cite{mcclintock_fabrication_2021} and auxetic \cite{lipton2018handedness} metamaterial body structures, and the active control of body stiffness \cite{ijspeert_central_2008} for successful locomotion in cluttered terrain.
In relation to this, we observed significant slip during locomotion, highlighting the need for effective ground interactions for successful thrust generation. 
We hope to incorporate passive \cite{autumn_frictional_2006, kim_smooth_2008} and active \cite{de_rivaz_inverted_2018} adhesion mechanisms in the next generation of CLARI.
Such controlled shape modulations could have significant implications for terradynamic streamlining \cite{li_terradynamically_2015} in cluttered terrains by changing the relative potential energy landscapes of surrounding complex environments \cite{othayoth_locomotor_2021}. 

We also observed that CLARI is unable to effectively utilize its entire bandwidth of leg cycling frequencies (up to the transmission resonances around \SI{40}{\hertz}) due to unfavorable body dynamics. 
As the next steps, we plan to characterize this in detail \cite{goldberg_gait_2017} as a function of both body compliance and shape in order to generate ideal leg trajectories and gaits for effective locomotion \cite{doshi_effective_2019}.
Another source of unwanted dynamics is the electrical tether for delivering power and control signals. 
Despite the current tethers being ultralight, their small movements seemed to perturb the motion of the robot causing it to drift laterally or turn. 
We expect that integrating power \cite{mcdonnell_enabling_2022} onboard the robot would not only enhance its autonomy but also make it more resistant to undesired perturbations.
Similarly, the robot leg modules were not identical in their performance despite our best efforts to match them, and more importantly, the articulated joints degraded during the process of robot testing. 
We expect improvements to the fabrication process and potentially small design changes (such as strengthening the structure around the SFB, etc.) will improve the robot operation lifetime and enable CLARI to be deployed in real-world environments.

Ultimately, with the above improvements incorporated, we hope that future generations of CLARI will be able to autonomously locomote through complex and cluttered natural environments and begin to deliver on the promise of significant socio-economic impact utilizing miniature robots.

\medskip
\textbf{Acknowledgements} \par %delete if not applicable))
We thank all members of the Animal Inspired Movement and Robotics Lab (AIM-RL) at the University of Colorado Boulder for invaluable support and discussions. We also thank Stephen Uhlhorn and 6DLaser for femtosecond laser micromachine development and fabrication assistance.

Any opinions, findings, conclusions or recommendations expressed in this material are those of the authors(s) and do not necessarily reflect the views of any funding agency. This work is partially funded through grants from the Paul M. Rady Mechanical Engineering Department, the US Army research office (ARO) Grant \# W911NF-23-1-0039 and the Meta Foundation (K.J.).

\textbf{Conflict of Interest Statement} \par %delete if not applicable))
The authors have no competing or conflicts of interest.

% References
\medskip

% Use the following code if you wish to generate your bibliography with BibTeX;
% replace the string "MSP-template" below with the name(s) of
% the BibTeX data base(s) you want to use.
% The resulting bibliography-output (the content of the .bbl file)
% must be pasted back into this file before submission.
% Please also include your BibTeX data base file(s) in your submission
% so that we can re-run BibTeX if necessary.
%
\bibliographystyle{MSP}
\bibliography{CLARI}

\clearpage

\begin{figure}
  \includegraphics[width=40mm]{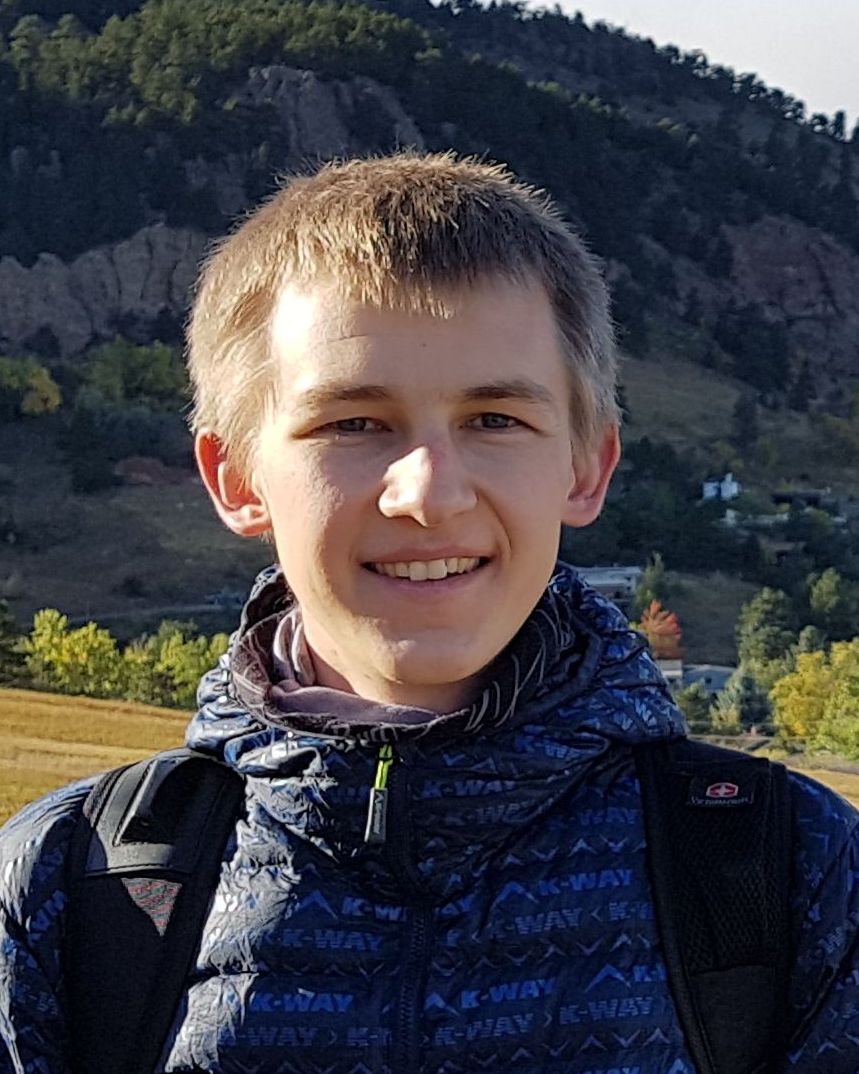}
  \caption*{Heiko Kabutz received his B.Eng. in Mechanical Engineering from the University of Pretoria, South Africa, in 2019. He received his M.S. in 2022 and is currently pursuing his Ph.D. in the Department of Mechanical Engineering at the University of Colorado Boulder, USA. His research interests include the mechanical design, biomechanics, insect-scale manufacturing, and control of legged movement mechanisms for bioinspired robotics.}
\end{figure}

\begin{figure}
  \includegraphics[width=40mm]{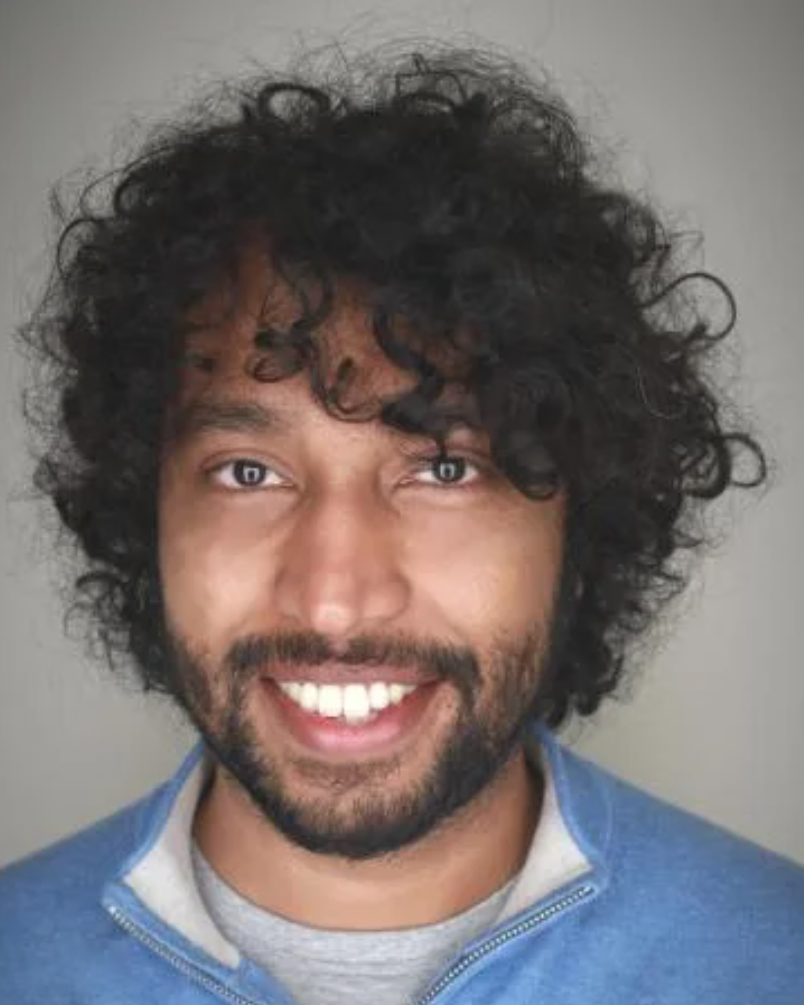}
  \caption*{Kaushik Jayaram received his B.Tech and M.Tech from Indian Institute of Technology Bombay in 2009 and his Ph.D. from University of Berkeley in 2015. He is currently an Assistant Professor in Robotics at the Paul M. Rady Department of Mechanical Engineering, University of Colorado Boulder and directs the Animal Inspired Movement and Robotics Laboratory (AIMRL). The groups research interests include bioinspired robotics, biomechanics, locomotion robustness, origami-based design and fabrication, and distributed sensing.}
\end{figure}

% Table of contents entry should be 50 - 60 words long
% Image should be 55 mm broad and 50 mm high or 110 mm broad and 20 mm high

\clearpage

\begin{figure}
\textbf{Table of Contents}\\
\medskip
  \includegraphics[width=0.8\linewidth]{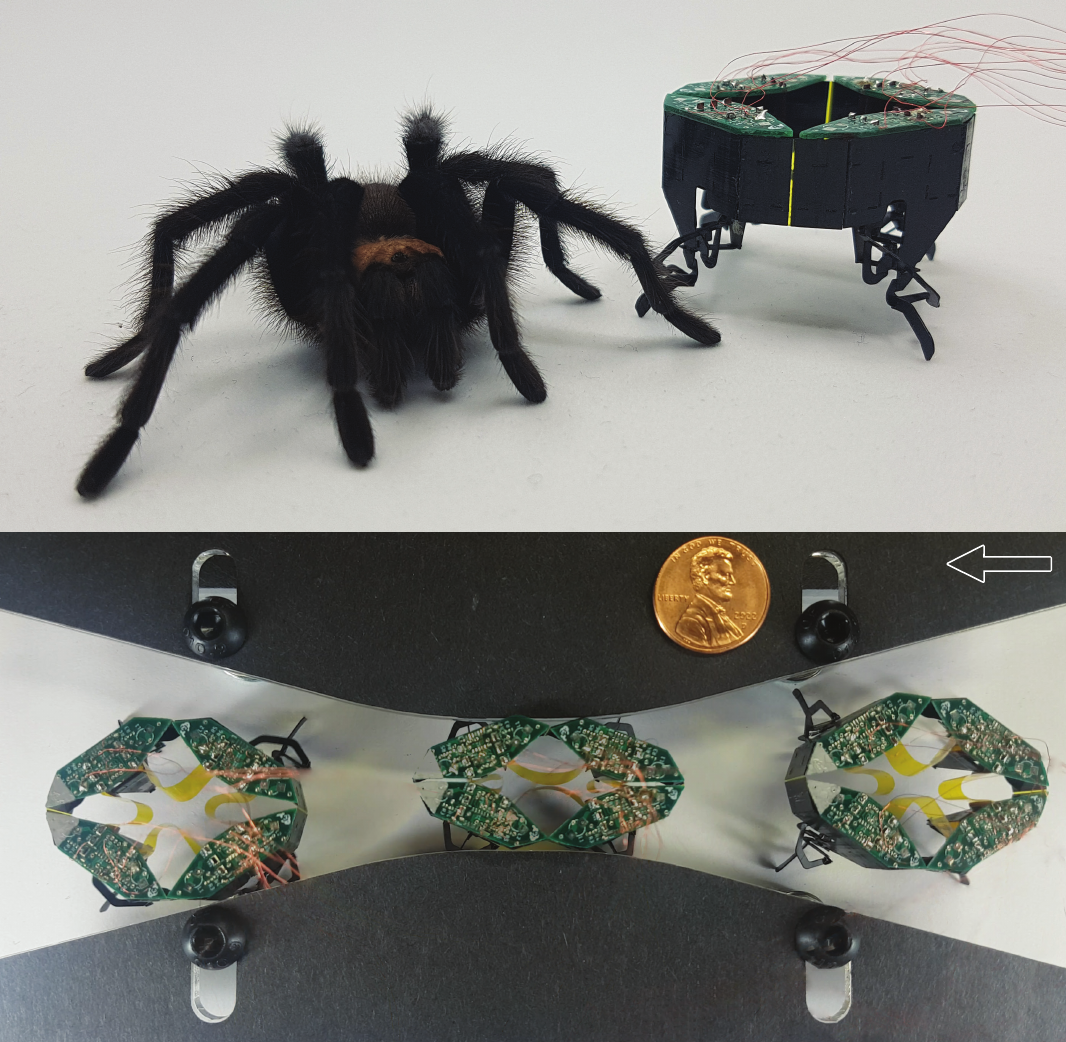}
  \medskip
  \caption*{We present CLARI — Compliant Legged Articulated Robotic Insect - an insect-scale (\SI{3.5}{\centi \meter}, \SI{2.59}{\gram}) modular origami shape-morphing quadrupedal robot. CLARI leverages its articulated exoskeleton for passive body shape adaptation and successfully locomotion through laterally confined environments - a first for legged robots.}
\end{figure}

\clearpage
\setcounter{page}{1}
{\title{Supporting Information}
\maketitle
}

\section*{Supplementary Text}

In the following sections, we provide additional details related to the manufacturing of CLARI and its subcomponents to expand on the summary in the main paper.

\subsection*{Fabrication Details}

The first step of the laminate fabrication process \cite{sreetharan_monolithic_2012, wood_microrobot_2008, whitney_pop-up_2011} is etching thin film materials based on kirigami designs. 
The primary machine tool used for this purpose is a custom built (6D-Laser) femtosecond laser (Light Conversion Carbide-CB5, with \SI{343}{\nano m} UV harmonics) micromachine. 
The system features a 6-axis processing surface (\SI{20}{\centi m}$\times$\SI{20}{\centi m} ablation area) controlled by sub-micron precision motion stages (ALIO 6-D Hybrid Hexapod) and a macro zoom camera (Bassler Inc) for part alignment. 
The laser paths are controlled by a galvo positioning system (ScanLab excelliSCAN) with a spot size of \SI{8}{\micro m} and a field of view area of about \SI{20}{\milli m}$\times$\SI{20}{\milli m} via custom software (Direct Machine Control, DMC Pro).
Ideal ablation parameters for the various materials (as noted below) were determined by cutting each material separately at a range of varying laser beam fluence values, cutting speeds and number of passes, followed by characterization of etch rate, laser kerf, and visible signs of material damage under a confocal microscope (Keyance VHX-7000). 
Following laser ablation, all parts were placed into an ethanol sonicator (Vevor) bath, cleaning the cut edges and surface debris off the parts, thus ensuring strong bonding while avoiding unintended electrical conductivity.

Once all individual materials were cut, we created custom laminates by alternately stacking structural materials with adhesives in a precision pin-alignment jig. 
For the lamination process that followed this step, a pneumatic heated press (MasterPress MP-20) was used. 
Custom heat and pressure curing profiles were programmed following manufacturer guidelines on material specifications.
Lamination was followed by a final laser ablation step to release the individual components and assembled using folding techniques.

\subsection*{Actuator Fabrication}
\label{sec:fab_act}
CLARI actuators were fabricated following the streamlined manufacturing process described by Jafferis \etal \cite{jafferis_streamlined_2021}, which overcomes the need for in-plane alignment of heterogeneous materials \cite{jafferis_multilayer_2016} and thus simplified fabrication and improved process yield and overall device performance. 
We used piezoceramics (PZT, T105-H4NO-2929 from Piezo Inc.) of \SI{125}{\micro m} thickness and coated with nickel electrodes as our active outer actuator layers. 
These were laser cut with a fluence of \SI{1.86}{\frac{J}{cm^2}} and a cutting speed of \SI{500}{\milli m s^{-1}}. 
To ensure a clean cut through the thickness of the material, 6 parallel lines spaced at \SI{5}{\micro m} from each other were used. 
To raster off the electrodes at the base and tip of the actuator, we used a double \SI{5}{\micro m} cross line hatch pattern and ablation fluence of \SI{0.38}{\frac{J}{cm^2}}.
The central (structural) actuator layer was uncured unidirectional carbon fiber (M55J/RS-3C) of \SI{50}{\micro m} thickness.
A hole of diameter \SI{0.8}{mm} was cut into the top piezoelectric layer of each actuator to allow access to the central conductive layer to transmit the drive signal.
Silver paint (16062, PELCO) was added into each actuator hole to electrically connect the center layer to the copper traces on the outside.
To bond solderable \SI{20}{\micro m} copper (Mitsumi Inc.) electrodes to the outer surface, we used a \SI{7.5}{\micro m} thick conductive adhesive (ESP8680-SC, AI Technology) cut at fluence of \SI{1.48}{\frac{J}{cm^2}}.
The complete actuator stack was heat pressed for 4h at \SI{170}{\celsius}.
As a structural reinforcement to minimize transmission losses, an additional support layer of carbon laminate (\SI{100}{\micro m} thick, 0-90-0) was bonded to the base of the actuator on either side with acrylic adhesive (FR1500, DuPont).

\subsection*{Actuator Testing Setup and Analysis}
\label{sec:test_act}
In order to drive the actuators, we used a unipolar alternating three-wire configuration \cite{karpelson_driving_2012} that includes a high voltage DC bias, a high voltage AC signal and ground. 
These required signals were generated off-board through a high voltage power amplifier (BD300, PiezoDrive, \SI{100}{\times}) connected to a remote real-time computer (IO135, Speedgoat) which produced custom analog voltages of \SI{0}{\volt} to \SI{3.3}{\volt}. 
The high-voltage signals were then routed to the actuators through ultrathin magnet wire (MWS wires). 
The remote computer also controlled the camera triggering and synchronized recording of force measurements via MATLAB Simulink Real-Time Toolbox.

For kinematic analysis, the actuators were mounted in a custom jig (inset, Figure \ref{fig:Actuator}c) and high speed (\SI{>10}{\times} actuation frequency) videos (Phantom V720, Vision Inc.) were collected and subsequently tracked using a trained neural network using DeepLabCut \cite{mathis_deeplabcut_2018}. 
Blocked force measurements were obtained using a 20g capacity, miniature load cell (LSB200, Futek).
The load cell was aligned to the actuator (barely touching) with a \SI{200}{\micro m} overlap during these experiments. 
All further data analysis was done with Python using custom scripts.

\subsection*{Single Leg Module Fabrication}
\label{sec:fab_leg}

The leg modules are composed of the frame, the transmission, actuator side walls, power-PCB and the leg in addition to the actuators described above.

The frame, transmission and leg are a single laminate with two \SI{90}{\micro m} thick $0-90-0$ carbon fiber structural layers sandwiching a \SI{25}{\micro m} polyimide (Kapton HN50, Dupont) as flexure layer, all bonded together by \SI{12}{\micro m} acrylic adhesives (FR1500, Dupont). 
The foldable assembly joints were designed to be \SI{200}{\micro m} and \SI{400}{\micro m} long for $\ang{90}$ and $\ang{180}$ folds respectively.
All motion joints were castellated and had an effective length of \SI{30}{\micro m} to allow for joint bending without interfering with the structural carbon fiber material thickness on each side.
The leg transmission was folded manually under a microscope with tweezers, using droplets of cyanoacrylate glue to secure folds.
The alignment of segments was ensured through the internal tabs and the use of a custom assembly jigs.

The side walls are composed of \SI{125}{\micro m} thick copper-clad fiber glass (Digikey) and help integrate the actuators into the \SI{400}{\micro m} power-PCB (custom fabricated through PCBWay) and transmission frame. 
The sidewalls were laser patterned to selectively ablate copper while drawing traces for routing power from the power-PCB to the actuators.

\subsection*{Single Leg Module Assembly}
\label{sec:ass_leg}
The assembly of the modular leg consisted of following steps. 
First, the actuators were slotted into the sidewalls and electrically connected with conductive paint (16062, PELCO) to create a rigid actuator sub-assembly.
Next, this sub-module was slotted into the transmission and the linkage tabs were folded into the actuator tip slots. 
Following this, the power-PCB was soldered onto the top of the actuator sub-assembly and aligned with the transmission ground. 
Wires to the external connects completed the leg module assembly.
Finally, all mechanical connections were further reinforced using droplets of cyanoacrylate glue.

\subsection*{Single Leg Module Testing}
The setup used for characterizing single leg modules is depicted in Figure \ref{fig:Leg_Experimental_Setup} below.

\subsection*{Whole Robot Assembly}
\label{sec:ass_rob}

The final component needed to be fabricated for the assembly of the robot was the single flexure body interconnecting the side walls. 
These side walls were identical to the body laminates except that we chose \SI{12.5}{\micro m} polyimide with flexure length designed to be \SI{300}{\micro m} to maximize body compliance.
The four individual legs were then connected together through the side walls interlocking to horizontal ribs within each module.

The complete distribution of the robot mass is shown in Table \ref{tab:MassDistribution}.
The mechanical robot skeleton, excluding the power-PCB, has a mass of \SI{1.812}{g} (comparable to similarly sized HAMR \cite{doshi_effective_2019} which weighs \SI{1.43}{g}), and more than half of this are the piezoelectric actuators.
As noted before (Section \ref{sec:des_modleg}), the power-PCB in the current CLARI is merely a key component required for the assembly of the leg module, but does not enable any power autonomy currently and therefore, serves as a payload for this robot. 
We anticipate future iterations of the robot will feature thinner versions of the power-PCB but with electrical components integrated for providing onboard power.

\subsection*{Laterally Confined Locomotion Setup}

To test the robot's ability to move through confined spaces, we constructed a custom arena with acrylic plates (\SI{6.3}{\milli \meter} thick). These plates were laser cut to the desired geometry and mounted on an optical table with standoffs which permitted the legs to freely swing under while constraining only the body.

\clearpage
\section*{Supplementary Tables}

\textbf{Table S1}: Mass distribution of CLARI components
\renewcommand{\thetable}{S1}
\begin{table}[!htbp]
    \centering
    \caption{Mass distribution of CLARI components}
    \begin{tabular}{c|c|c|c}
        \hline
        \textbf{Component} & \textbf{\#} & \textbf{Individual} & \textbf{Cumulative} \\
         &  & \textbf{Mass (mg)} &  \textbf{Mass (mg)} \\
        \hline
        Transmission & 4 & 113 & 452 \\
        Legs & 4 & 4 & 16 \\
        Side walls & 8 & 16 & 128 \\
        Actuators & 8 & 124 & 992 \\
        Interconnects & 8 & 28 & 224 \\
        Power-PCB & 4 & 195 & 780 \\
        \hline
        \textit{Mechanical subtotal} & & & \textit{1812}\\
        \textit{Electrical/Payload} & & & \textit{780}\\
        \hline
        \textbf{Total mass} & & & \textbf{2592} \\
        \hline
    \end{tabular}
    \label{tab:MassDistribution}
\end{table}

\clearpage
\section*{Supplementary Figures}

\textbf{Figure S1:} Experimental setup used to characterize the performance of the leg modules and actuators.
\renewcommand{\thefigure}{S1}
\begin{figure} [!htbp]
	\centering
	\includegraphics[width=0.5\linewidth]{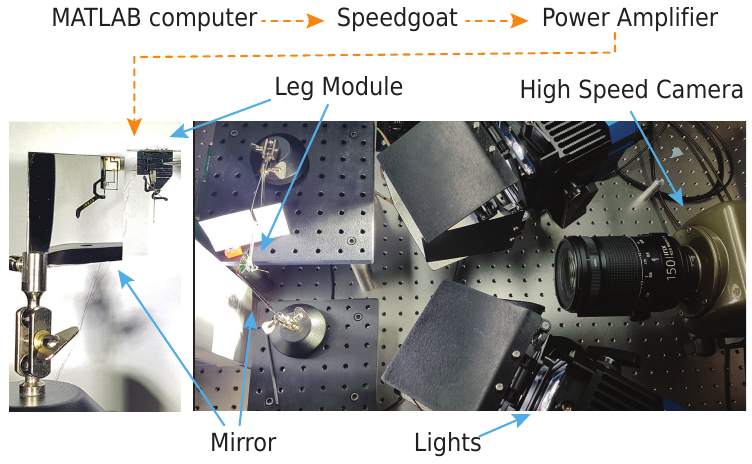}
	\caption{Experimental setup used to characterize the performance of the leg modules (and actuators) featuring high speed cameras for kinematic analysis and force transducer for blocked force measurements. An off-board computer delivers control signals to the power amplifier which is then routed through ultrathin cables.}
	\label{fig:Leg_Experimental_Setup}
\end{figure}

\textbf{Figure S2:} CLARI gait diagram.
\renewcommand{\thefigure}{S2}
\begin{figure} [!htbp]
	\centering
	\includegraphics[width=0.3\linewidth]{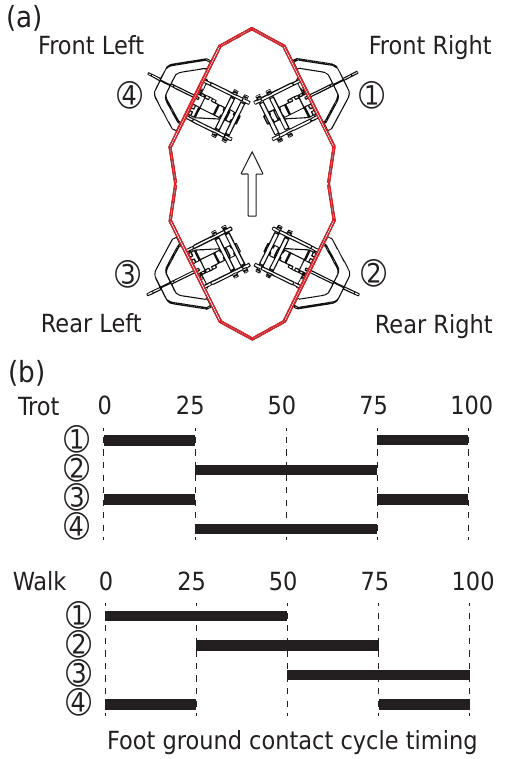}
	\caption{(a) Leg numbering and reference layout. (b) Foot ground contact timing for the trot and walk gait as a percentage of gait cycle.}
	\label{fig:Leg_trajectories}
\end{figure}

\textbf{Figure S3:} CLARI turning.
\renewcommand{\thefigure}{S3}
\begin{figure} [!htbp]
	\centering
	\includegraphics[width=0.7\linewidth]{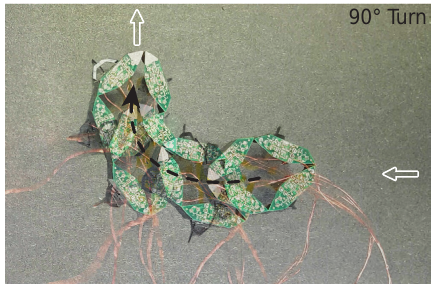}
	\caption{Robot performing \SI{90}{\degree} right turn with at 5Hz trot gait with leg module voltage signal biasing.}
	\label{fig:CLARI_turning}
\end{figure}

\clearpage
\section*{Supplementary Videos}
\textbf{Video S1:} Video summarizing the design, fabrication of CLARI and highlighting its locomotion performance

\end{document}